\documentclass[conference]{IEEEtran}
\IEEEoverridecommandlockouts
\usepackage{cite}
\usepackage{amsmath,amssymb,amsfonts}
\usepackage{algorithmic}
\usepackage{graphicx}
\usepackage{textcomp}
\usepackage{xcolor}
\usepackage{amsmath}
\usepackage[style=base]{caption}
\usepackage[utf8]{inputenc}
\usepackage{gensymb}
\usepackage[font=small,labelfont=bf]{caption} 

\definecolor{cb_red}{rgb}{0.89,0.1,0.11}
\definecolor{brown}{rgb}{0.59, 0.29, 0.0}

\def\BibTeX{{\rm B\kern-.05em{\sc i\kern-.025em b}\kern-.08em
    T\kern-.1667em\lower.7ex\hbox{E}\kern-.125emX}}
\begin{document}
\pagenumbering{arabic}
\title{UET-Headpose: A sensor-based top-view head pose dataset\\
}

\makeatletter
\newcommand{\linebreakand}{%
  \end{@IEEEauthorhalign}
  \hfill\mbox{}\par
  \mbox{}\hfill\begin{@IEEEauthorhalign}
}
\makeatother

\author{\IEEEauthorblockN{Linh Nguyen Viet\IEEEauthorrefmark{1}
\IEEEauthorblockA{UET AILab, VNU\\
Hanoi, Vietnam}
\IEEEcompsocitemizethanks{\IEEEcompsocthanksitem\IEEEauthorrefmark{1}equal contribution}
\IEEEcompsocitemizethanks{\IEEEcompsocthanksitem\IEEEauthorrefmark{2}corresponding author}
\and
\IEEEauthorblockN{Tuan Nguyen Dinh\IEEEauthorrefmark{1}}
\IEEEauthorblockA{UET AILab, VNU\\
Hanoi, Vietnam}
\and
\IEEEauthorblockN{Duc Tran Minh}
\IEEEauthorblockA{UET AILab, VNU\\
Hanoi, Vietnam}
\linebreakand
\IEEEauthorblockN{Hoang Nguyen Viet}
\IEEEauthorblockA{UET AILab, VNU\\
Hanoi, Vietnam}
\and
\IEEEauthorblockN{Quoc Long Tran\IEEEauthorrefmark{2}}
\IEEEauthorblockA{UET SISLAB, VNU\\
Hanoi, Vietnam}}
}

\maketitle

\begin{abstract}

Head pose estimation is a challenging task that aims to solve problems related to predicting three dimensions vector, that serves for many applications in human-robot interaction or customer behavior. Previous researches have proposed some precise methods for collecting head pose data. But those methods require either expensive devices like depth cameras or complex laboratory environment setup. In this research, we introduce a new approach with efficient cost and easy setup to collecting head pose images, namely UET-Headpose dataset, with top-view head pose data. This method uses an absolute orientation sensor instead of Depth cameras to be set up quickly and small cost but still ensure good results. Through experiments, our dataset has been shown the difference between its distribution and available dataset like CMU Panoptic Dataset \cite{CMU}. Besides using the UET-Headpose dataset and other head pose datasets, we also introduce the full-range model called FSANet-Wide, which significantly outperforms head pose estimation results by the UET-Headpose dataset, especially on top-view images. Also, this model is very lightweight and takes small size images.
\end{abstract}


\section{Introduction}
Today with the strong development of technology infrastructure, security cameras are widely used to observe and monitor behavior. From this massive amount of data obtained, it is beneficial to extract the necessary information. Many research addresses the head pose estimation problem, which has numerous applications, such as identifying customer behavior and driver behavior monitoring. Most head pose estimation methods use the 300W-LP  dataset or the BIWI dataset, which just has estimated head pose in small range yaw angles $[-99, 99]$. So some methods use another dataset - CMU Panoptic Dataset \cite{CMU} to have full-range yaw [-179, 179]. However, the CMU Panoptic Dataset dataset has the disadvantage that the background of the images is similar because it was recorded in the same setup. Moreover, this laboratory environment is hard to set up, requiring multiple cameras with high resolution to attach around the room. We have tons of public cameras; our proposed methods will not be convenient and adaptive for using available cameras.
This paper also developed FSANet-Wide, which extends head-pose estimation to the full range of yaws (hence wide) by improving FSANet \cite{FSANet} architecture. In doing so, we make the following contributions:
\begin{itemize}
    \item We introduce a new data-driven approach to collecting head pose images based on an IMU sensor, various devices and a top-view camera.
    \item We also publish the new dataset, namely UET-Headpose collected by new method.
\end{itemize}
 
\par 

\section{Related works}

\textbf{Head pose estimation}. 
In 2018, the FAN model \cite{FAN} was announced, that work by detected facial landmarks, the MAE evaluation results reached $7.89$ on the BIWI \cite{BIWI} dataset and $9.12$ on the AFLW2000 \cite{300WLP} dataset.\par
The method based on landmark points also has a considerable disadvantage: When the face orientation is completely greater than $60^{\circ}$, it will be complicated to find these landmarks. Therefore, direct head direction estimation models from the image without detecting the landmarks were carried from this fatal drawback.\par
In 2018, Ruiz and colleagues proposed the HopeNet model \cite{Hopenet}, a convolutional neural network using multiple loss functions trained to predict Euler angles directly from the image. The classification loss also is used to guides the networks to learn the neighborhood of poses robustly. The evaluation results with the HopeNet \cite{Hopenet} model on the BIWI \cite{BIWI} dataset reached $4.9$ and were state-of-the-art when published.\par
In 2019, Yang and colleagues published the FSA-Net \cite{FSANet} model and became the best model at the time of publication when evaluated on the BIWI \cite{BIWI} dataset with an MAE of $4.00$, and the model was very lightweight at only $5.1$MB. The FSA-Net \cite{FSANet} model also only needs to use a single RGB image without depth information or video. The vital improvement of the FSA-Net \cite{FSANet} plot in two parts is feature extraction, using a phased regression model (SSR) to predict face orientation with a light size and attention mechanism.\par
The Rankpose \cite{RankPose} and QuatNet \cite{QuatNet} models both use the ResNet50 \cite{Resnet} backbone to estimate head poses. The improvements of these models are mainly in the loss function. QuatNet model \cite{QuatNet} estimates predict the quaternion of head pose instead of Euler angles. QuatNet \cite{QuatNet} uses two regression functions, L2 regression loss, and ordinal regression loss, to address the non-stationarity property in head pose estimation by ranking different intervals of angles with the classification loss. The resulting model also gives good results with an MAE of $4.13$ with the BIWI \cite{BIWI} dataset.\par

\textbf{Head pose dataset}. Currently, there are two primary datasets for training: 300W-LP \cite{300WLP} and BIWI \cite{BIWI}, corresponding two main dataset for testing: AFLW2000 \cite{300WLP} and a part of BIWI \cite{BIWI}.\par
Two datasets 300W-LP \cite{300WLP} and AFLW2000 \cite{BIWI}, are created by 3DDFA \cite{300WLP}. This method passes RGB images through morphable 3D and outputs another RGB image with head pose changed. The 300W-LP \cite{300WLP} dataset includes $122450$ images, and AFLW-2000 \cite{300WLP} includes $2000$ images. In two datasets, yaw angles in the range of $[-90,90]$, pitch and roll angles in the range $[-99, 99]$.\par
BIWI \cite{BIWI} dataset includes $24$ videos recorded by Kinect. People in this dataset are sitting in front of the camera and turning their heads in different directions. Head poses are estimated by in-depth information from Kinect. BIWI \cite{BIWI} dataset includes over $15000$ images; the head pose range covers about $[-75, +75]$ degrees of yaw angle and $[-60, +60]$ degrees of pitch angle.\par
All three datasets 300W-LP \cite{300WLP}, AFLW-2000 \cite{300WLP}, and BIWI \cite{BIWI}, have been used by many models for training and evaluation of the FSA-Net \cite{FSANet}, Rankpose \cite{RankPose}, and achieved good results when the mean error (MAE) was only about $4^{\circ}$. \par
However, a major drawback of the 300W-LP \cite{300WLP}, AFLW-2000 \cite{300WLP}, and BIWI \cite{BIWI} datasets is that the head pose angles (including yaw, pitch, roll) are all in the range of $[-89, 89]$ degrees and these data are collected from depth camera to create 3D head orientation, so the prediction model is less effective with large-angle data such as from security cameras.\par 
From that disadvantage, WHENet \cite{WHENet} used additional head pose datasets extracted from the CMU Panoptic dataset \cite{CMU}. Head pose dataset from CMU Panoptic dataset \cite{CMU} providing comprehensive yaw angle face orientation from $[-179, 179]$ improved a lot when evaluated with camera data compared to models trained only with 300W-LP \cite{300WLP} dataset.\par
Another problem is evaluating models, in older datasets like BIWI \cite{BIWI}, AFLW-2000 \cite{300WLP}, there are not enough angle to evaluate full range, yaw model. From that disadvantage, the paper further developed a dataset UET-Headpose with images captured directly from a system that combines a camera and an angle sensor mounted directly on the face. It is very adaptive; the device and set up everywhere quickly can collect for the actual situation.\par

\begin{figure*}[ht]
    \centering
    \includegraphics[width=0.85\textwidth]{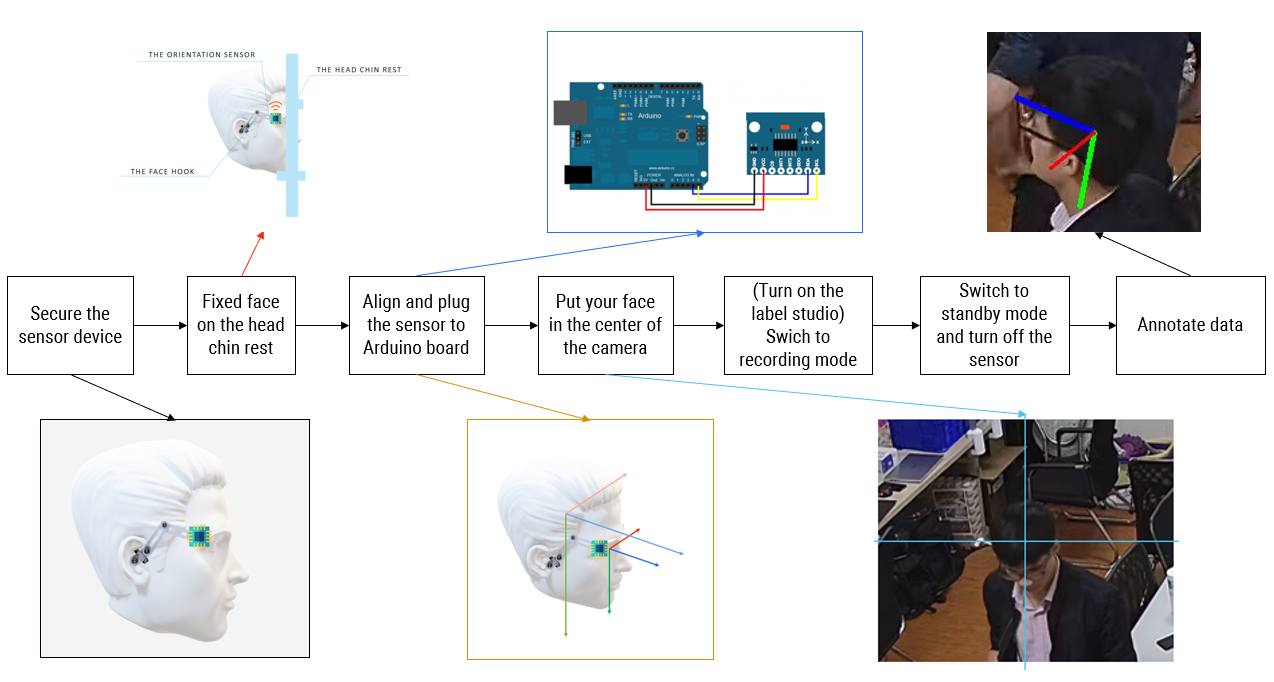}
    \captionsetup{justification=centering}
    \caption{The detailed architecture system of new approach.}
    \label{fig:systemoverall}
\end{figure*}

\section{Proposed Method}

In this chapter, we will present in detail the approach to solving the above problem. First, in section 3.1, we will introduce the new approach for the face orientation data acquisition, the overall architecture of the new approach in Fig. \ref{fig:systemoverall}, we have seven main phases, with the beginning is the process of wearing sensor for humans, and we have annotated data in the final phase. Then, section 3.2 will analyze the new dataset and present a baseline model for head pose estimation $360^{\circ}$ for security cameras. A new model is called FSANet-Wide, and using this model with a new dataset UET-Headpose-train and two other datasets.\par

\textbf{Structure Design}
We propose a new dataset called UET-Headpose with the advantages of background richness, overcoming the disadvantages of the head pose dataset extracted from the CMU Panoptic dataset \cite{CMU} while ensuring the distribution of yaw angle is uniform in the range $[-179, 179]$ degrees. Moreover, our proposed setup is very efficient for carrying in real life.\par
The UET-Headpose dataset was created to capture the head pose of annotated people in many conditions with a wide yaw range and top view security cameras. The system includes a head-mounted sensor module, an arduino aboard, a surveillance camera, a chin rest for fixing the head, and a server to control, store and process data. In this paper, the surveillance camera used is HIK Vision 4 MP ColorVu Fixed Bullet Network Camera (DS-2CD2T47G1-L).\par
The surveillance camera is placed high with an inclination angle as zero-point compared to the fixed face $62.458$, $-35.601$ and $-8.096$ for roll, pitch, yaw angle, respectively. This result is found by experimentally measuring the deflection angle between the plane parallel to a chin rest and the surveillance camera through the sensor. Firstly, the sensor is fixed to a plane of the rectangular box and is guaranteed to be perpendicular to the four surrounding planes. Then, the rectangular box is placed on a plane parallel to the support in the yaw, pitch, roll direction.\par
The chin rest is attached directly to the edge of the table. The design of the chin rest \cite{HeadChinRest} includes 02 table legs, 02 fixed pivot bars, 01 adjustable height forehead support bar, 01 adjustable height chin rest bar, 01 adjustable depth chin rest, and 02 fixed bars sensor. Notably, the two fixed axles use 3030 aluminum profiles. The rest of the parts are 3D printed in 105 hours. The phase \textit{``Fixed face on the head"} in Fig. \ref{fig:systemoverall} shows the chin rest and the way to locating the head of people on that.\par
The head-mounted sensor module equipment for recording the head's angle value includes the 9-axis smart sensor MCU-055, the Arduino board WeMos D1 R2, a 10000 mAh battery, an ear hook, and wires. The device is designed to ensure optimal speed and increase certainty to limit sensor slippage and signal loss.\par
The system also includes one management computer for storage and processing data.\par

\textbf{Transmit angle data from board to server}

The sensor is connected to the Arduino board Wemos D1 R2 with the I2C protocol. The advantages of the I2C protocol simplicity and high reliability. The I2C protocol uses a clock line (SCL) transmitted at 100kHz and 400kHz, two wires GND and VCC, are responsible for stabilizing and supplying current to the sensor.
The Arduino board Wemos D1 R2 is used due to the built-in 2.4 GHz Wifi router. The data from the sensor will be transmitted to the Arduino board continuously with a maximum frequency of 100 Mhz. The Arduino board captures data and transmits it to the computer via the MQTT \cite{MQTT} protocol.\par
The Arduino circuit has been directly integrated with the wifi router so that sensor information will be transmitted to the management computer via wifi to the MQTT \cite{MQTT} broker built on the management computer.\par
The Arduino circuit receives the signal from the sensor through 4 wires SCL, SDA, GND, and VCC. This information has been completely processed at the sensor, so the Arduino board will only receive and read yaw angles, pitch angles, and roll angles from the sensor and send it to the MQTT \cite{MQTT} broker.\par
Every $0.04$s, the Arduino board will send data to the computer once in a JSON file, including information fields: time, x, y, z, where time is the time to record data from the sensor, obtained directly through the server \textit{pool.ntp.org}. Three values x, y, z correspond to 3 information yaw, pitch, roll returned by the sensor. \par
The computer that stores and processes the data will continuously receive information from the sensor but selectively record it when the manager chooses. Based on the uploaded JSON file timer, the computer will request the camera the corresponding frame. We use this camera to provide images with a resolution of $2688$ × $1520$, 60Hz: $25$fps, 4mm focal length.\par
These frames will be manually labeled before being fed into the final training model. The labeling tool that we use is called ``Eva." \cite{EVA} The advantage of this tool is that it is light, simple to deploy, and use because it is packaged into a complete docker.\par

\textbf{Sensors}

We use an angle sensor to record poses. Many types of sensors are used on the market that can be combined with an Arduino board, such as GY-50, MPU-6050, MPU-9250, MCU-055, ...
Sensor MCU-055 is a sensor developed by Bosch. In addition to integrating nine axes like the MPU-9250 sensor, sensor MCU-055 integrates a microprocessor responsible for combining parameters from the sensor, gyro, accelerometer, and magnetic field direction sensor. It returns absolute angular direction (Eulerian vector, 100Hz sampling frequency) as triaxial data based on the 360-degree sphere through a 32-bit microcontroller running BSX3.0 FusionLib software developed by Bosch. The raw data has been wholly preprocessed with algorithms and gives the final stable angular direction result.\par

\textbf{Design board for wearing on head}

The extended board consists of main parts: a sensor, an ear hook, an Arduino board, a battery, and all kinds of wires.\par
The sensor will be fixed through the face hook through a soft steel wire to align the sensor to each face, but still, make sure to keep the sensor fixed after alignment without tilting or deviating. After experimenting with many different types of wire, the soft steel wire used to fix the face hook is 1mm non-steel wire. Larger steel wires will be, but adjusting the sensor angle will be difficult, and conversely with smaller wires will not guarantee to hold the sensor firmly. However, the face hook version no. 1 using steel wire has the disadvantage of fine-tuning the sensor due to the elastic force of the steel wire. So the second version is made by 3D printing method was developed with the advantages: possible to adjust the directions through 3 rotating axes, no elastic phenomenon like steel wire. The face hook version no. 1 includes a soft steel wire responsible for fixing the sensor and ear hook. The sensor will be fixed to align the sensor to each face, but still, keep the sensor fixed after alignment without tilting or deviating. The image in block \textit{``Secure the sensor device"} of Fig.\ref{fig:systemoverall} shows the way the sensor is attached on the head of a person.\par
The ear hook is responsible for attaching tightly to the ear and keeping it fixed to the sensor. The first version of the hook is made entirely of soft silicone imported from abroad, it has the advantage of being soft and light. However, the ear attachment is not fixed because it has not yet adhered to the ear. Therefore, a 2mm steel thread was used by parallelizing the ear hooks through the soft black adhesive tape to overcome this shortcoming. With this additional steel wire, each time the sensor is attached, it can be easily adjusted to suit each person's ears. When adding a steel wire, the ear hook was firmly attached around the ear, but it was still slightly shaken when tested.\par
This vibration error is identified because the upper end of the ear hook is not firmly attached to the fulcrum in the ear. So a foam cotton pad using PE Foam material was used. This sponge will tightly cushion the inside of the ear combined with the previously existing thin rim to form a solid structure, minimizing the vibration of the sensor.\par

\textbf{Server}

The management computer is used for annotating data will have the following tasks: Create an MQTT \cite{MQTT} broker to act as an intermediary between the sensor and the Arduino board, a system for receiving and storing, and processing data, an Eva labeling tool. The pipeline head pose and image data synchronization for the server is illustrated in Fig.\ref{fig:pipelineserver}\par
The data receiving and processing system is responsible for storing the packets that the MQTT \cite{MQTT} broker requires the manager. The frames corresponding to the timestamps in the packet returned by the server and storing the data included raw images, head pose labels, and face position labels.\par
The data acquisition system is coded using python language, including two separate processing threads for collecting data from sensors and managing received data. The receiver stream will run continuously $24/24$ as soon as the system is turned on. The received data management flow will be displayed as a terminal, switching between two data recording modes and non-data recording.\par
When the system receives information from the MQTT \cite{MQTT} broker as a sensor result (JSON file containing time, yaw, pitch, roll fields) by default, the information will not be stored. However, when the data storage computer administrator switches to information storage mode, the system records information from the sensor to a log file. Each line corresponds to the information from the JSON file.\par
After having a log file containing information about the face orientation of the labeled person, the system will automatically retrieve the corresponding frames through the camera's API. Since the surveillance camera only provides an API to playback saved video, view live video, and get the live snapshot, the method used to retrieve frames is to get a datum of the video start frame ( corresponding to the first frame in the log file), get the video from that point. Then, the time of the following frames will be calculated based on the FPS and the sequence number of that frame. The camera used has an FPS frequency of $25$ frames per second, so the number of packets sent by the sensor per second is chosen to be $25$, so there will be a difference between the frame time and the packet time. Therefore, the message is a maximum interval of $0.04$ seconds. 
\begin{figure}
    \centering
    \captionsetup{justification=centering}
    \includegraphics[width=0.47\textwidth]{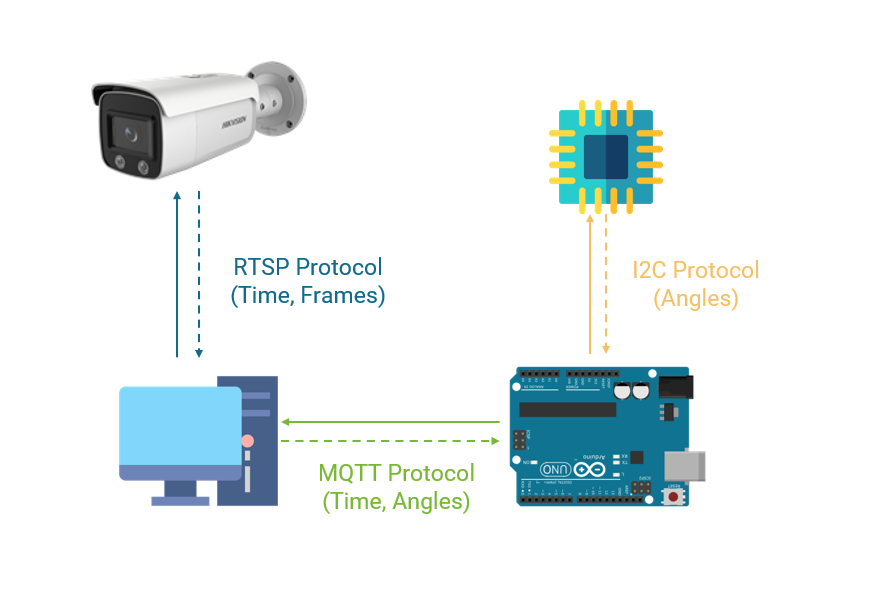}
    \caption{The pipeline of pose data and image data synchronization}
    \label{fig:pipelineserver}
\end{figure}

\textbf{Cost advantage}

The method of collecting BIWI dataset and CMU panoptic dataset requires investment in at least one 3D camera \cite{BIWI}, or ten 3D cameras, 480 VGA videos, 31 HD videos \cite{CMU}, so the cost of implementing this data acquisition system can be very expensive. It can be up to US\$5000 accompanied by a complicated installation and deployment process.\par

Our face-to-face receivers are designed for simplicity and ease of deployment, including components: one sensor, one small 3d-printed system, the ear hook, and a 2D camera. These are very available components and easily obtainable for less than US\$150. Easy to deploy and install anywhere with a variety of environments. In Fig.\ref{fig:compare}, we describe differences between our proposed dataset and available dataset.\par


\subsection{Analyze UET-Headpose dataset and baseline model}

After collection, the UET-Headpose dataset includes $12.848$ images obtained from $9$ people with the distribution of yaw angles as shown in Fig. \ref{fig:distribution}. So, the UET-Headpose dataset will be divided into seven people for training with $10.848$ images and two people for evaluation with $2.000$ images. These images will be cropped to the face on a scale increased by $40\%$ vertically and $60\%$ horizontally compared to the face. Label files also have additional coordinate information (x-min, y-min, x-max, y-max) so that the face can be accurately cut from the cropped image.\par
The UET-Headpose dataset had a uniform yaw angle distribution for all directions from $[-179, 179]$. The dataset is obtained by having the annotated people rotated all yaw directions when collecting the dataset. So the dataset will make it possible for the model to learn all yaw angles within a $360^{\circ}$ range. Fig.\ref{fig:distribution} illustrates a more extensive coverage of the yaw angle of our proposed dataset, the yaw angle of our proposed dataset covers a wider range than 300W-LP and also fixes the drawbacks of the CMU-Headpose dataset.\par

\begin{figure}
    \centering
    \captionsetup{justification=centering}
    \includegraphics[width=0.47\textwidth]{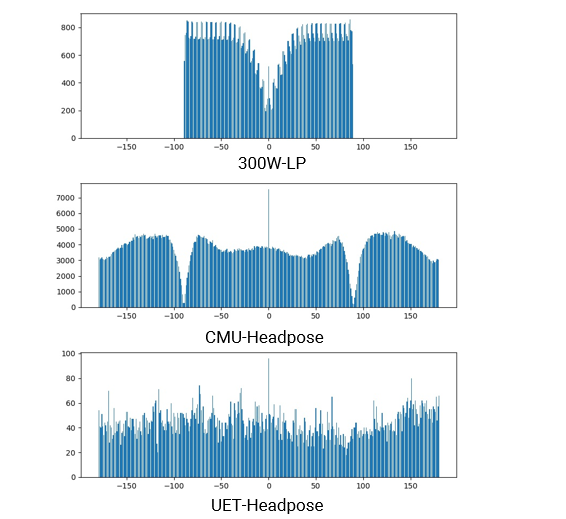}
    \caption{Comparison between CMU, 300W-LP and UET-Headpose distribution of the yaw angle}
    \label{fig:distribution}
\end{figure}

\textbf{Model FSANet-Wide}

FSA-Net \cite{FSANet} model was introduced by Yang et al. at the 2019 CVPR conference. This model has the advantage of being a very lightweight model with only $1.2$ million parameters for combining three sub-models, which gives good results when the average error with each angle is only $4^{\circ}$. Therefore, the baseline will be based on the FSA-Net \cite{FSANet} model, doing experiments and improving it.\par

Original FSANet \cite{FSANet} model achieves good results in 300W-LP \cite{300WLP}, as described above, the 300W-LP \cite{300WLP} dataset has yaw pose in a range $[-99, 99]$. So this model results in the yaw pose range in $[-99, 99]$. We will improve the SSR module in FSANet \cite{FSANet} and fusion mechanism by expand range of output model in a full-range yaw pose $[-179, 179]$. 

\section{Experiments}
In this section, the experiments will show the results and the evaluation of the proposed models. Before going into the detailed evaluation of the models, we will outline how to conduct the experiment and the evaluation methods. Specifically, Section 4.1 describes the data. Section 4.2 details the evaluation method. Section 4.3 shows result of the models mentioned in section 3.
\subsection*{4.1 Dataset}
\subsubsection*{1. 300W-LP \cite{300WLP} dataset} \label{300W-LP}
The 300W-LP \cite{300WLP} dataset includes $122450$ images, with a yaw angle in the range $[-89, +89]$.  All images are saved in .jpg image format; the head pose labels are saved in the .mat file format.
\subsubsection*{2. Head pose dataset extracted from CMU Panoptic dataset \cite{CMU}} 
The original CMU Panoptic dataset \cite{CMU} consists of $83$ sequences. Each sequence includes a JSON file of camera parameters and videos corresponding to each camera, a file of face coordinates in 3D space, a file of coordinate points of 3D human parts, $480$ VGA quality videos, and $31$ HD quality videos.\par
After processing the CMU Panoptic dataset \cite{CMU} for the head pose problem, a dataset contains $1342018$ images. The yaw angle distribution of the dataset is almost the same as the uniform distribution, at angles near $90$ and $-90$ will be less due to the effect of gimbal lock. Two angles pitch and roll, the magnitudes are in the range $[-89, 89]$. Details of the processing steps are presented in \cite{WHENet}.
\subsubsection*{3. Our new dataset UET-Headpose}
The results are obtained from a new dataset consisting of $12,848$ images. Then, $2,000$ images were randomly divided into the UET-Headpose-val dataset and $10,484$ images into the UET-Headpose-train dataset. The data will be processed and converted to HDF5 format to optimize training time. \par
All datasets will be converted to HDF5 format with the help of the h5py library. Our experiment showed it expertly necessary. A model trained with data in jpg format needs more than 8x times longer HDF5 data.\par

\textbf{Experiment setup} \par
In order to conduct training and evaluation of the model, we use the 300W-LP \cite{300WLP} dataset and Head pose data extracted from the CMU Panoptic dataset \cite{CMU} and the UET-Headpose-train dataset for training. The AFLW2000 \cite{300WLP} and UET-Headpose-val are used for testing. \par
We used Pytorch for implementing the proposed FSA-Net \cite{FSANet}. For data augmentation in training, we applied numerous filters in the albumentation library to training images. We choose the MSE function as the loss function for the training; MAE and MAWE for evaluation. To train the models, we used the ADAM optimizer. Batch size is selected as $128$, and the learning rate is initialized as $0.001$. The learning rate was reduced by a factor of $0.1$ every $30$ epochs.\par

\subsection*{4.2 Metric evaluation}

Normally, pose estimation methods are trained with the 300W-LP \cite{300WLP} dataset. The mean absolute error (MAE) is used as the evaluation metric. However, in this scenario, the head pose is estimated in the wide-range pose (yaw angles). This evaluation method will not be reasonable for the problem of evaluating wide-angle direction up to $360^{\circ}$. For example, when the actual angle is $170^{\circ}$, the predicted angle is $-170^{\circ}$; then, the two angles are only $40^{\circ}$ apart, but the MAE value calculated is $340^{\circ}$. So, in the evaluation with UET-Headpose-val, we use MAWE \cite{WHENet} instead: 

$$ MAE\ =\ \frac{1}{n}\sum_{i=0}^{n}(|{y_i\ -\ \hat{y}}_i|) $$

$$ MAWE = \frac{1}{n}\sum_{i=0}^{n}min(|\theta_{pred} - \theta_{true}|, 360 - |\theta_{pred} - \theta_{true}|) $$

\begin{table}[t]
    \captionsetup{type=figure, position=above, justification=centering}
    \captionof{table}{Evaluation models in the AFLW2000 dataset} \label{tab:num-inter}
\begin{center}
\begin{tabular}{lcccc}
\hline
\multicolumn{1}{|l|}{\textbf{Model}}            & \multicolumn{1}{l|}{\textbf{Yaw}}  & \multicolumn{1}{l|}{\textbf{Pitch}} & \multicolumn{1}{l|}{\textbf{Roll}} & \multicolumn{1}{l|}{\textbf{MAWE}} \\ \hline
\multicolumn{1}{|l|}{FSA-Net \cite{FSANet} (Fusion)} & \multicolumn{1}{c|}{4.5}           & \multicolumn{1}{c|}{6.08}           & \multicolumn{1}{c|}{4.64}          & \multicolumn{1}{c|}{5.07}          \\ \hline
\multicolumn{1}{|l|}{WHENet \cite{WHENet}}         & \multicolumn{1}{c|}{5.11}          & \multicolumn{1}{c|}{6.24}           & \multicolumn{1}{c|}{4.92}          & \multicolumn{1}{c|}{5.42}          \\ \hline

\multicolumn{1}{|l|}{FSA-Net-Wide (1x1)}        & \multicolumn{1}{c|}{5.75}          & \multicolumn{1}{c|}{5.98}           & \multicolumn{1}{c|}{3.07}          & \multicolumn{1}{c|}{4.93}          \\ \hline
\multicolumn{1}{|l|}{FSA-Net-Wide (var)}        & \multicolumn{1}{c|}{4.36}          & \multicolumn{1}{c|}{5.95}           & \multicolumn{1}{c|}{3.27}          & \multicolumn{1}{c|}{4.52}          \\ \hline
\multicolumn{1}{|l|}{FSA-Net-Wide (w/o)}        & \multicolumn{1}{c|}{5.01}          & \multicolumn{1}{c|}{6.32}           & \multicolumn{1}{c|}{3.28}          & \multicolumn{1}{c|}{4.87}          \\ \hline
\multicolumn{1}{|l|}{FSA-Net-Wide (Fusion)}     & \multicolumn{1}{c|}{4.59}          & \multicolumn{1}{c|}{5.69}           & \multicolumn{1}{c|}{\textbf{2.85}} & \multicolumn{1}{c|}{\textbf{4.37}}          \\ \hline
\multicolumn{1}{c}{}                            & \multicolumn{1}{l}{}               &                                     & \textbf{}                          &                                   
\end{tabular}   
\label{tab:evaluate_AFLW2000}
\end{center}
\end{table}

\subsection*{4.3 Results}

\begin{table*}[ht]
\centering
\captionsetup{justification=centering}
\caption{Evaluation models in two datasets AFLW2000 and UET-Headpose-Val}
\begin{tabular}{|l|c|c|c|c|}
\hline
\multicolumn{1}{|c|}{\textbf{Models}}                                             & \textbf{\begin{tabular}[c]{@{}c@{}}Full-range\\ yaw\end{tabular}} & \textbf{\begin{tabular}[c]{@{}c@{}}Model\\ size\end{tabular}} & \textbf{\begin{tabular}[c]{@{}c@{}}MAWE \\ UET-Headpose-Val\end{tabular}} & \textbf{\begin{tabular}[c]{@{}c@{}}MAWE\\ AFLW2000\end{tabular}} \\ \hline
FSA-Net \cite{FSANet} (Fusion)                                                          & No                                                             & 2.91 Mb                                                               & 54.03                                                                    & 5.07                                                             \\ \hline

WHENet \cite{WHENet}                                                                    & \textbf{Yes}                                                       & 17.1 Mb                                                               & 53.65                                                                    & 5.42                                                             \\ \hline
\begin{tabular}[c]{@{}l@{}}FSA-Net-Wide (Fusion)\\ 300W-LP\end{tabular}            & No                                                             & 2.91 Mb                                                               & 52.76                                                                    & \textbf{4.37}                                                             \\ \hline
\begin{tabular}[c]{@{}l@{}}FSA-Net-Wide (Fusion)\\ 300W-LP + CMU\end{tabular}      & \textbf{Yes}                                                       & 2.91 Mb                                                               & 56.72                                                                    & 10.77                                                            \\ \hline
\begin{tabular}[c]{@{}l@{}}FSA-Net-Wide (Fusion)\\ UET-Headpose-train\end{tabular} & \textbf{Yes}                                                       & 2.91 Mb                                                               & 9.3                                                                      & 35.89                                                            \\ \hline
\begin{tabular}[c]{@{}l@{}}FSA-Net-Wide (Fusion)\\ All datasets\end{tabular}       & \textbf{Yes}                                                       & 2.91 Mb                                                               & \textbf{7.29}                                                            & 7.55                                                             \\ \hline
\end{tabular}
\label{tab:HA-result}
\end{table*}

In this section, we will show our baseline when training with different combinations of dataset.\par
Table \ref{tab:HA-result} summarizes key results from FSANet-Wide-fusion in four type data training. We compare with model size, mean average wrapped errors on AFLW-2000 \cite{300WLP}, UET-Headpose-Val, and report full-range MAE. 
From this table, we can see that because our proposed dataset has different distribution with other datasets, the available state-of-the-art model will not achieve high accuracy in our dataset. \par

\textbf{Full-range results and comparisons} \par 

From Table \ref{tab:HA-result}, FSA-Net with fusion mechanism, WHENet\cite{WHENet}  achieve and FSA-Net-Wide after training with 300W-LP very bad performance in our dataset, with $54.03$, $53.65$ and $52.76$ in MAWE, respectively. But our baseline model, FSA-Net-Wide trained with 300W-LP achieves the highest performance at MAWE of AFLW2000, with $4.37$. After using our proposed dataset, we obtained the baseline trained with all dataset to get the lowest MAWE on our validation dataset, the figure is $7.29$.

\textbf{Narrow range results and comparisons} \par 

The experimental result in Table \ref{tab:evaluate_AFLW2000} shows the performance of WHENet and FSA-net in different mechanisms, including our proposed baseline on AFLW2000. FSA-Net-Wide with fusion mechanism shows best performance in this dataset with $4.37$ MAWE overall, it is noticeable that the MAWE on Roll angle of this model is significantly small, the figure is $2.85$.\par

\section{CONCLUSION}
We have proposed the new approach for collecting a head pose dataset with easy to setup on any cameras and low-cost, so this approach can keep data distribution near real problems. And we also provide our dataset acquired through the proposed approach and baseline model for this dataset. In the future, we wish to extend our current dataset to many situations and apply it to handle the real-life problem.
\section*{Acknowledgment}
This work has been supported by VNU University of Engineering and Technology. 


\end{document}